\pdfoutput=1

\documentclass[11pt]{article}

\usepackage{naacl2021}

\usepackage{times}
\usepackage{latexsym}

\usepackage[T1]{fontenc}

\usepackage[utf8]{inputenc}

\usepackage{microtype}
\usepackage{graphicx}
\usepackage{subfigure}
\usepackage{booktabs} 
\usepackage{array}
\usepackage{amssymb}
\usepackage{amsmath}
\usepackage{nicefrac,xfrac}
\usepackage{listings}
\usepackage{placeins}
\usepackage{enumitem}
\usepackage{tabulary}
\usepackage{xcolor}

\title{Ad Headline Generation using Self-Critical Masked Language Model}

\author{Yashal Shakti Kanungo \\
  \texttt{yashalk@amazon.com} \\\And
  Sumit Negi \\
  \texttt{suminegi@amazon.com} \\\And
  Aruna Rajan \\
  \texttt{rajarna@amazon.com} \\}

\begin{document}
\maketitle
\begin{abstract}
For any E-commerce website it is a nontrivial problem to build enduring advertisements that attract shoppers. It is hard to pass the creative quality bar of the website, especially at a large scale. We thus propose a programmatic solution to generate product advertising headlines using retail content. We propose a state of the art application of Reinforcement Learning (RL) Policy gradient methods on Transformer \cite{vaswani_attention_2017} based Masked Language Models \cite{devlin_bert_2019}. Our method creates the advertising headline by jointly conditioning on multiple products that a seller wishes to advertise. We demonstrate that our method outperforms existing Transformer and LSTM + RL methods in overlap metrics and quality audits. We also show that our model-generated headlines outperform human submitted headlines in terms of both grammar and creative quality as determined by audits.
\end{abstract}

\section{Introduction}

\begin{figure*}[]
	\vskip 0.2in
	\begin{center}
		\centerline{\includegraphics[width=\linewidth]{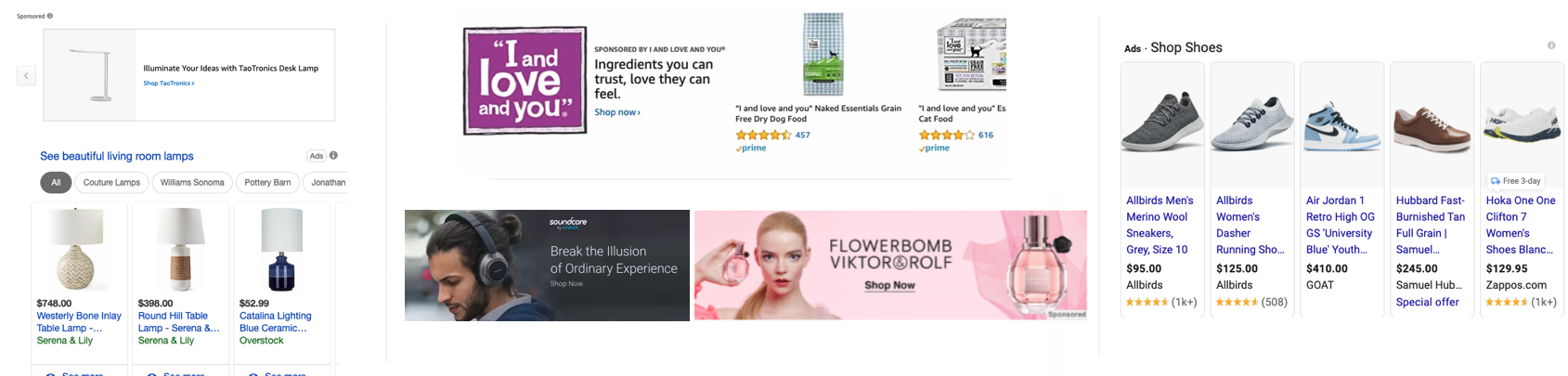}}
		\caption{Examples of different product ads from multiple websites across the internet. A variety of ad headlines accompany the products in these ads.}
		\label{fig:sb_example_wide}
	\end{center}
	\vskip -0.2in
\end{figure*}

There are a various types of ads. 
A set of example ads that showcase products selected by sellers along with headlines that advertise them are shown in Figure \ref{fig:sb_example_wide}. Sellers create multiple ad campaigns for multiple products, bid in an auction to advertise and pay for clicks on the ad.

An E-Commerce product catalog may have millions of products which can be advertised. To ease the ad headline writing process, humans resort to programmatically padding keywords, or repasting the retail catalog content in the advertisement.


Templated creatives such as ``Save Now on ..." or ``Buy more (product) of (brand)" save the creative effort but fail to create any excitement or brand identity in the minds of shoppers. High quality headlines are more attractive to shoppers and offer better value proposition. In this paper, we describe how we built a Natural Language Generation (NLG) system to generate instantaneous, attractive and brand identity building headlines for advertisements that intend to promote a wide range of products offered by a brand. 



The content associated with a retail product has challenging characteristics. Some product titles have poor structure, grammatical issues, or partial phrases. The product titles also include varying number of product features such as \textit{``Hyper Tough 18V Cordless Drill, 3/8 inch Chuck, Variable Speed, with 1.2Ah Nickel Cadmium Battery, Charger, Bit Holder  LED Light" } along with titles such as \textit{``ZIPIT Grillz Backpack, Camo Grey"}.


The generated headlines need to capture the information present in the retail attributes and at the same time be different and uniquely attractive. Advertisers select multiple related products that are advertised as part of a single ad campaign. The ad campaign headline is then shared across all of these related products. Thus, the headline also needs to generalize the shared characteristics of the products and cannot be specific to a single product within the campaign.

The key contributions of our work are:

\begin{itemize}
	\item We use Masked Language Model (MLM) for the generation of advertisement headlines using multiple products at the same time. Extensive test-set metrics, quality and grammar audits show that the proposed model outperforms all the baselines and the human-submitted headlines in terms of quality and grammar.
	\item The novel usage of RL for the training of MLM allows us to directly optimize the MLM for improved headline quality metrics without changing inference setup or latency. Our method can also be applied to any other NLG task such as summarization, translation etc.
	\item Our model reduces the extensive effort and time that is required to manually create headlines and has low latency.
\end{itemize}

\section{Related Work}

Natural Language Understanding (NLU) using Language Models (LM) has observed great leaps in recent years. LMs have evolved from using word level models \cite{joulin_bag_2016} to 
to a variety of extensions to the Transformer \cite{vaswani_attention_2017}.
The BERT \cite{devlin_bert_2019} employs Transformer in a pre-training setting and introduced the MLM training objective.

 \citet{ramachandran_unsupervised_2016} first demonstrated textual generation by using auto-regressive prediction in a seq2seq architecture. Transformer based auto-regressive methods such as GPT2 \cite{radford_language_2019} and BART \cite{lewis_bart_2019} which predict one word at a time have also shown good results. \citet{zhu_incorporating_2020} concatenated BERT representations with the Encoder and Decoder layers of another LM to incorporate pre-trained LM.  Another model \cite{dong_unified_2019} combines BERT-based Transformer Encoder with attention masking from the Transformer decoder. \citet{rothe_leveraging_2019} combined pre-trained BERT Encoder with GPT decoder for NLG. 


\citet{ranzato_sequence_2016} framed NLG as an RL problem and the generation quality as a reward.  The Self-Critical Sequence Training (SCST) approach \cite{rennie_self-critical_2017} replaces the learned baseline from other approaches  \cite{bahdanau_actor-critic_2017} with the model's own inference time algorithm to normalize the rewards.


For advertising, recent works \cite{xu_clickbait_2019,hughes_generating_2019} have combined LSTM based pointer network \cite{see_get_2017} with RL methods to generate advertisement headlines. While these methods improve the results, they fail to utilize extensive pre-training of Transformer based models and their various well-demonstrated advantages.

Our method extends BERT based generation \cite{dong_unified_2019} by using Self-Critical policy gradient method \cite{rennie_self-critical_2017} and jointly conditioning the generated sentence on multiple products at the same time. This allows us to use pre-trained BERT based LMs that can be trained to optimize various inference time metrics that are typically non-differentiable such as BLEU, Rouge, Readability etc.

\section{Self-Critical Masked Language Model}

\subsection{Masked Language Model}
\label{section:mlm_base}
The BERT model takes an unlabeled input sequence $x = (x_1,x_2,...,x_{|x|})$ and randomly masks some positions $M_x$ by replacing them with a special mask token $\textnormal{[MASK]}$, to produce a sequence like $(x_1,\textnormal{[MASK]},...,x_{|x|})$. All the tokens are embedded and added to special positional embeddings. It then uses $N$ identical Transformer layers to generate contextualized representation, with each layer employing self-attention by taking in the output of the previous layer. To compute self-attention, the output of the previous layer is projected into triplets of vectors named Query, Key and Value ($Q, K, V$) of dimensions $d$. The attention $A$ is then given as:

\begin{equation} 
	\label{eq:mlm_attention}
	A=\textnormal{softmax}(\frac{QK^T}{\sqrt{d}})V
\end{equation}

After the final Transformer layer the model uses a feed forward layer  followed by a softmax over the vocabulary to predict the masked tokens. The MLM loss for the sequence $x$ is then calculated as:

\begin{equation}
	\label{eq:mlm_loglikelihood}
	\mathcal{L}_{MLM} = - \log \prod_{m \in M_x} p(x_m| (x_{m'} \in x \setminus M_x))
\end{equation}

where $(x_{m'} \in x \setminus M_x)$ represents all the tokens in $x$ that are not masked and $m \in M_x$ are all the masked positions.

\subsection{Encoding multiple products and common headline for Proposed MLM}

\label{section:proposed_mlm}


\begin{figure}[t]
	\vskip 0.2in
	\begin{center}
		\centerline{\includegraphics[width=\linewidth]{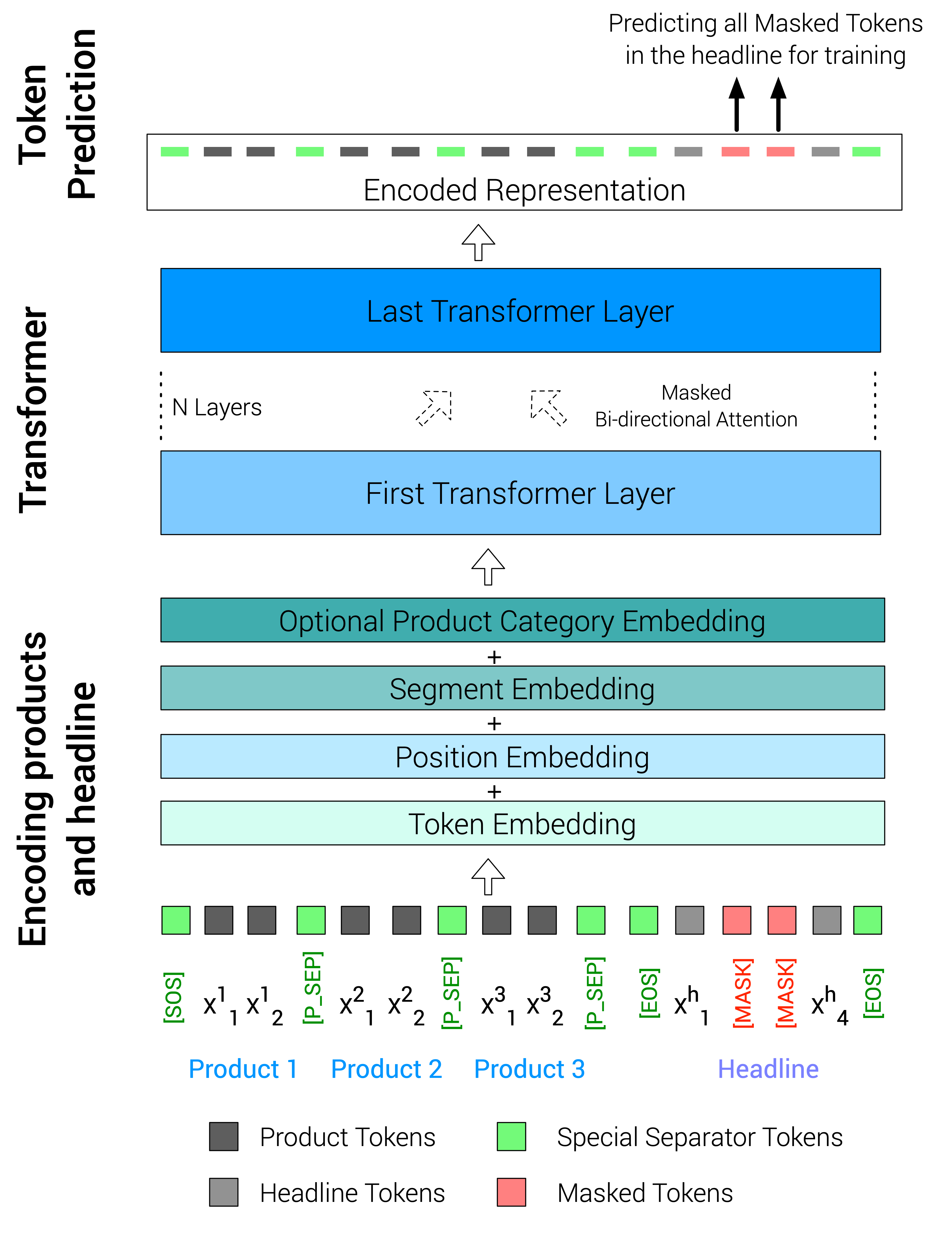}}
		\caption{The sub-tokens from the product titles and headline are embedded and added with other embeddings that encode the positional and segment information. We also optionally add an embedding that represents the category of the product. During training, the masked tokens are predicted using Transformer layers and the cross-entropy (Eq. \ref{eq:mlm_loglikelihood}) loss and Self-Critical (Eq. \ref{eq:rl_reward_scst_loss_approx}) gradient is used to optimize the model. During inference, we predict one word at a time (left-to-right) in an auto-regressive manner using Beam Search.}
		\label{fig:encoding}
	\end{center}
	\vskip -0.2in
\end{figure}


During training, for a given advertising campaign, our model takes as input it's headline $x^h = (x^h_1,...,x^h_{|x^h|})$ and a set $P$ of one or more products. Each product $p$ is represented by its title $x^p = (x^p_1,...,x^p_{|x^p|})$. The titles and the headline are tokenized to sub-word tokens.

To encode using the model that only accepts a single product, we simply append `\texttt{[EOS]}' $\in \mathbb{V}$ to both the title and the headline and concatenate their tokens. The entire concatenated sequence is prepended with `\texttt{[SOS]}' $\in \mathbb{V}$.

We encode multiple products by concatenating the tokens from different products using a special token `\texttt{[P\char`_SEP]}' $\in \mathbb{V}$. We replace a token `\texttt{[UNUSED\_0]}' $\in \mathbb{V}$ that remains unused during pre-training, with this special token during multi-product fine-tuning. This makes a distinction between different titles as well as the source and target sub-sequences. It also yields individual embeddings for each product for other tasks.

Only the tokens from the headline $x^h$ are randomly masked with token `\texttt{[MASK]}' $\in \mathbb{V}$. We discuss results for the model that additionally also masks the source tokens in section \ref{section:results_overlap}.

The complete process for an example such that all products in the ad have two tokens and the headline has 4 tokens is illustrated in Figure \ref{fig:encoding}. 


We also experimented with adding of category based embeddings. The category labels for each product such as ``Cell Phones and Accessories" are tokenized to subword units, encoded using the same embedding matrix as that of the title tokens, averaged and added to the title token embeddings.

\subsection{Generation using Self-Critical Masked Language Model}

\label{section:proposed_sc-mlm}

\begin{figure}[t]
	\vskip 0.2in
	\begin{center}
		\centerline{\includegraphics[width=\linewidth]{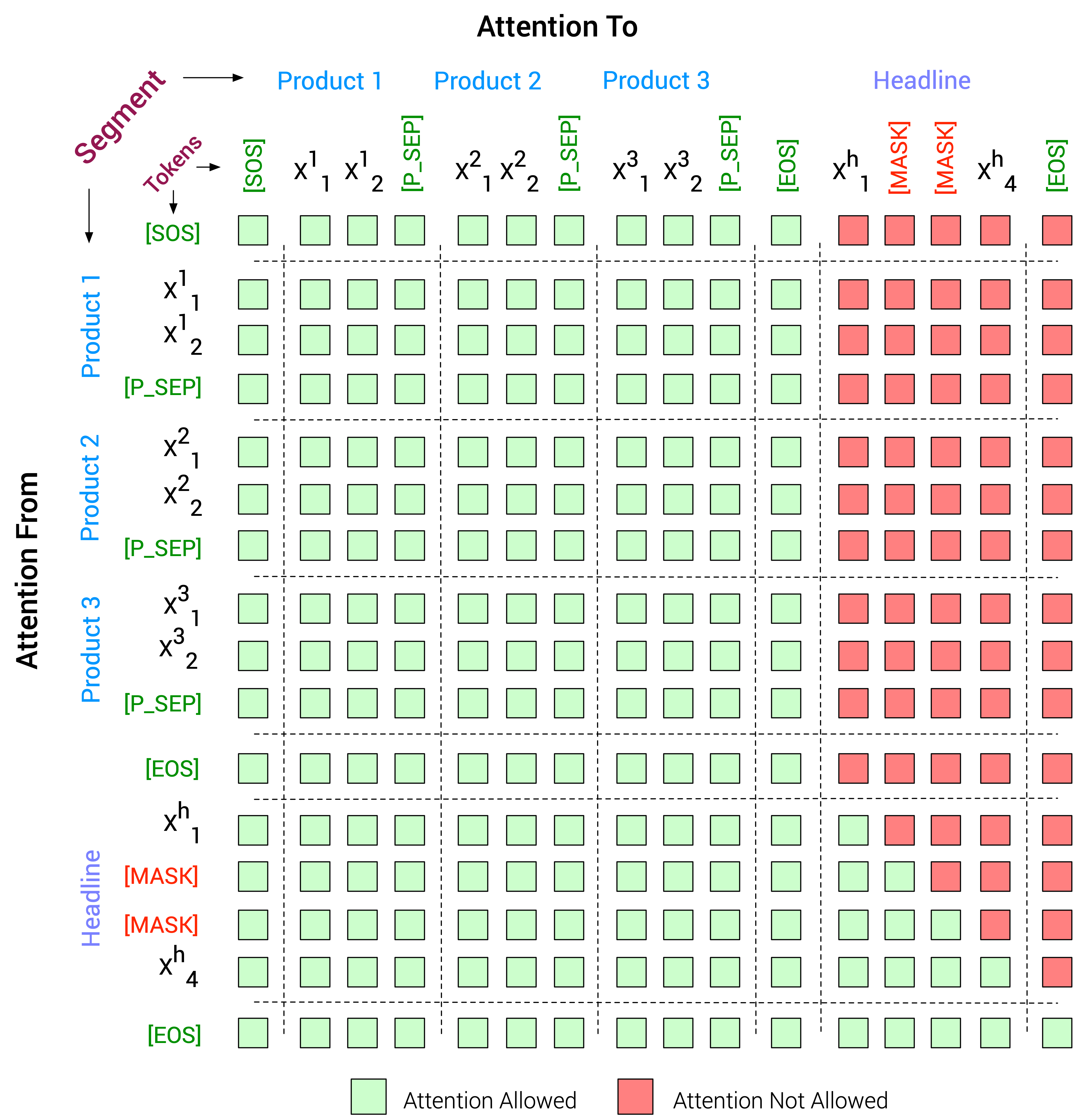}}
		\caption{Masked attention partially restricts attention for some token pairs. It prevents attention to headline tokens that would not be accessible during each step of generation during inference.}
		\label{fig:attention}
	\end{center}
	\vskip -0.2in
\end{figure}

The BERT MLM framework with multi-directional attention discussed in Section \ref{section:mlm_base} cannot be used for auto-regressive generation directly. This is because, during training, the masked headline words may condition on the future words which are not available during auto-regressive inference. For MLM auto-regressive generation, we employ masked attention \cite{dong_unified_2019} that modifies the attention from equation \ref{eq:mlm_attention} as below:

\begin{equation} 
	\label{eq:mlm_masked_attention}
	A_{\textit{masked}}=\textnormal{softmax}(\frac{QK^T}{\sqrt{d}} + \Phi_{ij})V
\end{equation}

where $\Phi_{ij}$ represents the attention mask between the positions $i$ and $j$. The elements are set to 0 if attention is allowed and $- \infty$ if it is not allowed. Figure \ref{fig:attention} illustrates the attention mask for headline generation using multiple input products. 

The BERT MLM uses log-likelihood (Equation \ref{eq:mlm_loglikelihood}) of masked words during training to optimize the model parameters. The likelihood is predicted using other ground-truth words during training and other predicted words during inference. This causes exposure bias \cite{ranzato_sequence_2016,rennie_self-critical_2017} and accumulates error during inference. Moreover, the training is optimized for log-likelihood, while we actually care about other more evolved measures of headline quality such as overlap metrics BLEU \cite{papineni_bleu_2002} and ROUGE \cite{lin_rouge_2004}.


To overcome these issues and improve the quality of the generated headlines, we frame the MLM as an RL problem. The model is an `agent' that takes the `action' of predicting masked words and updates the `state' such as the self-attention weights. The MLM follows a policy $\pi_\theta$ defined by the parameters $\theta$ of the model. It receives a reward that is proportional to the quality of the generated headline. This quality may either be the overlap with ground truth headlines that have been approved by internal subject-matter-experts or be predicted by another model. Our goal is to maximize the reward corresponding to a generated headline $\hat{x}^h$ during training, with the tokens at some masked positions $M_{x^h}$ sampled from the model.

We thus minimize the negative expected reward defined by any reward function $r(\cdot)$ for headline quality $r(\hat{x}^h)$ as:

\begin{equation}
	\label{eq:rl_reward_loss}
	\mathcal{L}_{RL} = - \mathbb{E}_{\hat{x}^h \sim \pi_\theta} [r(\hat{x}^h)]
\end{equation}

We can compute the gradient $\nabla_\theta \mathcal{L}_{RL}$ using the REINFORCE algorithm \cite{williams_simple_1992}. It is defined as:

\begin{equation}
	\label{eq:rl_reward_reinforce_loss}
	\nabla_\theta \mathcal{L}_{RL} = - \mathbb{E}_{\hat{x}^h \sim \pi_\theta} [r(\hat{x}^h) \nabla_\theta P]
\end{equation}

where,
\begin{equation}
	\label{eq:rl_probability}
	P = \sum_{m \in M_{\hat{x}^h}} \log p_\theta(\hat{x}^h_m| (\hat{x}^h_{m'} \in \hat{x}^h \setminus M_{\hat{x}^h})
\end{equation}

such that $M_{\hat{x}^h}$ are the masked positions and $\hat{x}^h \setminus M_{\hat{x}^h}$ are all the unmasked tokens.

To reduce the variance without changing the expected gradient, the algorithm proposes to use a baseline $b$ that does not depend on the generated headline $\hat{x}^h$. $b$ is used to normalize the reward along with $P$ from equation \ref{eq:rl_probability} as:

\begin{equation}
	\label{eq:rl_reward_reinforce_baseline_loss}
	\nabla_\theta \mathcal{L}_{RL} = - \mathbb{E}_{r(\hat{x}^h) \sim \pi_\theta} [(r(\hat{x}^h)-b) \nabla_\theta P]
\end{equation}

A single Monte-Carlo sample for each set of products and headline can be used to approximate the gradient. Using the definition of $P$ from equation \ref{eq:rl_probability}, we have the approximate gradient:

\begin{equation}
	\label{eq:rl_reward_reinforce_baseline_loss_approx}
	\nabla_\theta \mathcal{L}_{RL} \approx - (r(\hat{x}^h)-b) \nabla_\theta P
\end{equation}

Instead of using other models to estimate the expected baseline reward \cite{ranzato_sequence_2016,bahdanau_actor-critic_2017}, we employ Self-Critical training \cite{rennie_self-critical_2017} that involves generating two headlines using the same underlying MLM. The first headline $\hat{x}^h$ is generated by sampling from the vocabulary distributions generated by the model for the masked tokens. The second headline $\hat{z}^h$ is generated using the inference time strategy, which uses the token with the maximum probability at each step rather than sampling. The difference in the reward achieved by these two headlines is used to compute the gradient:

\begin{equation}
	\label{eq:rl_reward_scst_loss_approx}
	\nabla_\theta \mathcal{L}_{SC\_MLM} \approx - (r(\hat{x}^h)- r(\hat{z}^h)) \nabla_\theta P
\end{equation}

where $P$ is defined by equation \ref{eq:rl_probability}.

Thus, this method maximizes both the reward of the headlines generated by MLM and the likelihood of correct words by incorporating both the likelihood and the reward in the loss function.

\subsection{Inference}

During inference, we generate the headline auto-regressively using beam search until we reach the predetermined max length or each beam generates the end token. We have employed a modified version of Length Normalization \cite{wu_googles_2016} to better adapt to our headline lengths and training setup. This is necessary as the default beam search setup uses the log probability of each word to select the best headline. However, this biases the results as longer headlines would have lower probability of generation. We thus use the following normalized scores for each word to select the best headline:

\begin{equation}
	\label{equation:length_normalization}
	\textnormal{score}(\hat{x}^h_i) =  \textnormal{log-likelihood}(\hat{x}^h_i) * \frac{(2 + 1)^\alpha}{(2 + i)^\alpha}
\end{equation}

where $\alpha$ is the length normalization coefficient and $\hat{x}^h_i$ is the $i^{th}$ word of the generated headline in each beam. We also include additional Regular Expression based post-processing to remove extra spaces around various symbols such as `-,+()' etc. 

\section{Experiments}


\subsection{Training and Inference}
We used over 500,000 ad campaigns that were created on Amazon by sellers who have signed-up for advertising. Each campaign contains a set of related products along with an ad headline. We only selected the campaigns that contained English headlines and products with English titles. They were also de-duplicated to only have unique products-headline pairs. The mean product title length is 19.6
words and the mean headline length is 6.16 words.
The entire dataset was divided into train (85\%), validation (5\%) and test (10\%) sets. For training, we only selected the campaigns that comply with ad \citeauthor{amazon_advertising_policies_sponsored_link} as verified by internal experts.


We use HuggingFace \cite{wolf_huggingfaces_2020} implementation of Transformer BERT `Large' models as the base for our experiments. The models are pre-trained on WikiPedia and BookCorpus \cite{devlin_bert_2019,dong_unified_2019}. We first fine-tune the pre-trained model for up-to 15 epochs with early stopping using $\mathcal{L}_{MLM}$ and Adam \cite{adam}. We then further fine-tune the model for another 15 epochs with early stopping using Adam with $\nabla \mathcal{L}_{SC\_MLM}$ (Equation \ref{eq:rl_reward_scst_loss_approx}). We use the Rouge L F1 \cite{lin_rouge_2004} overlap with the approved headlines as the headline quality reward. For a fair comparison, the MLM-only model is fine-tuned for upto 30 epochs.

The model training is very time expensive with a single fine-tuning sub-experiment of 30 epochs taking over 20 days on an Nvidia v100. We thus only performed the essential experiments that help to determine the contribution of different sub-experiments and proposals. We estimated post-experiment that a single fine-tuning sub-experiment of 30 epochs would consume approximately 150 kWh of energy based on the GPU's power draw.


\subsection{Baseline}
We used a Pointer Network \cite{see_get_2017} based bi-LSTM with intra-decoder and temporal attention. We also used Self-Critical training with the bi-LSTM, similar to other ad headline generation methods \cite{xu_clickbait_2019,hughes_generating_2019} methods for a fair comparison to Self-Critical MLM.

\subsection{Ablations}
We trained a model with the same architecture, number of parameters and input as the proposed models but without MLM pre-training and separately without Self-Critical loss to study the impact of the proposals.

We also trained a model with MLM pre-training but fine-tuning only using the primary first product from each campaign instead of using all the products. This is interesting since some of the campaigns are cohesive to a degree with similar products and using only one product improves training time and inference latency.

We also report overlap metrics for model that does not use length normalization and post-processing discussed in equation \ref{equation:length_normalization}. We also include results for model that uses BERT Base as the base model instead of BERT Large.

\section{Results}

\subsection{Overlap with Approved Headlines}
\label{section:results_overlap}
The first evaluation criterion we adopt is overlap \cite{sharma_relevance_2017} of model headlines with subject-matter-experts approved human-submitted headlines from the test set (Table \ref{table:main_results}).

Masking the source product title words reduces the performance as the titles and headlines do not follow the same sentence structure and distribution. Adding product category embedding reduces performance and our hypothesis is that this is because the base model cannot be pre-trained with these embeddings. Only using one title achieves lesser but respectable performance, highlighting the efficacy of multi-product conditioning.

``No pre-training of MLM" highlights the advantage of using non-pretrained Transformer based architecture over bi-LSTM. `Proposed MLM' shows the advantage of using pre-training, BERT Large and only masking the headline. `Proposed Self-Critical MLM' achieves the best scores across all the metrics and highlights the applicability of our proposed approach.

\begin{table*}[!t]

	\vskip 0.15in
	\begin{center}
		\begin{small}

			\begin{tabular}{lccccc}
				
				\toprule
				Model & Rouge-L & CIDEr & BLEU-4 & METEOR & Avg. Cos. Sim. \\
				\midrule

				\textit{Baseline bi-LSTM Pointer Network model} \\
				
				\hspace{3mm} bi-LSTM  &  - & - & -  & - & - \\
				
				\hspace{3mm} Self Critical bi-LSTM  & 0.62 & 0.01 & 1.06 & 0.42 & -4.31 \\

				\midrule
				
				\multicolumn{4}{l}{\textit{MLM Baselines and Ablations (Single Product and No Self Critical Training)}}\\
				
				\hspace{3mm} First Product Only & 2.14 & 0.19 & 5.03 & 3.55 & 0.36 \\
				
				\hspace{3mm} First Product and Category embedding & 1.52 & 0.13 & 4.18 & 2.938 & 0.15 \\
				
				\midrule
				
				\multicolumn{4}{l}{\textit{Proposed MLM and Ablations (Multiple Products and No Self Critical Training)}}\\
				
				\hspace{3mm} Using BERT Base instead of BERT Large &2.85 & 0.22 & 4.96 & 3.58 & 1.53 \\
				
				\hspace{3mm} No pre-training of MLM (Training from scratch)& 3.38 & 0.27 & 5.72 & 3.79 & -0.04 \\
				
				\hspace{3mm} Additional Source Titles Masking & 4.13 & 0.29 & 4.42 & 5.41 & -2.09 \\
				
				\hspace{3mm} \textbf{Proposed MLM} & 5.08 & 0.42 & 7.49 & 5.46 & 1.31 \\
				
				\midrule
				
				\textit{Proposed Self-Critical MLM (SC-MLM) and Ablation}\\

				\hspace{3mm} No beam search normalization and post-processing & 5.37& 0.43 & 7.81 & 5.61 & 1.96 \\
				
				\hspace{3mm} \textbf{Proposed Self-Critical MLM} & \textbf{6.33} & \textbf{0.55} & \textbf{9.11} & \textbf{6.14} & \textbf{3.75} \\
				
				\bottomrule
			\end{tabular}
		\end{small}
	\end{center}
	\vskip -0.1in
	\caption{Absolute improvement over baseline in terms of overlap measures with over 50,000 manually approved human-submitted headlines from the test set. We have reported the\textbf{ differences} in the F1 of Rouge-L and BLEU-4 scores to the baseline bi-LSTM model. `Avg. Cos. Sim.' is the average cosine similarity of model headlines to the human-submitted headlines measured using an independently pre-trained Language Model.}
	\label{table:main_results}
\end{table*}

\subsection{Quality and Grammar Audits}

\begin{table*}[!t]
	\vskip 0.05in
	\begin{center}
		\begin{small}
			\begin{sc}
				\begin{tabular}{lcccc}
					\toprule
					& SC-biLSTM & MLM - Single Product & Proposed MLM & Proposed SC-MLM\\
					\midrule
					\multicolumn{5}{l}{\% improvement in mean rating over human-submitted headlines}\\
					\hspace{3mm}      &  -9.87\% & 0.40\% & 1.15\% & \textbf{2.07\%}      \\
					
					\midrule
					\multicolumn{4}{l}{\% improvement in number of headlines}\\
					
					\hspace{3mm} Rated $\geq 2$ out of 3    &  -4.99\% & \textbf{2.75\%} & 2.42\% & 2.37\% \\
					\hspace{3mm} Rated  3 out of 3  & -42.96\% & -0.06\% & 1.22\%  & \textbf{6.53\%} \\
					
					\bottomrule
				\end{tabular}
			\end{sc}
		\end{small}
	\end{center}
	\vskip -0.1in
	\caption{Comparison of model-generated headlines to human-submitted headlines on a 3-point scale quality audit of a random blind test set (N $\approx$ 5000).}
	\label{table:quality_scores}
\end{table*}

\begin{table*}[!t]
	\vskip 0.1in
	\begin{center}
		\begin{scriptsize}
			\renewcommand{\arraystretch}{0.92}
			\setlength{\tabcolsep}{4pt}
			\begin{tabular}{>{\centering\arraybackslash}m{0.26\linewidth} >{\centering\arraybackslash}m{0.222\linewidth} >{\centering\arraybackslash}m{0.222\linewidth} >{\centering\arraybackslash}m{0.222\linewidth}}
				\toprule
				\textbf{One of the source product's title} & \textbf{Human Submitted Headline} & \textbf{Proposed MLM} & \textbf{Proposed SC-MLM} \\
				\midrule
				BEST Natural Hair Growth Oil for GUARANTEED Hair Strength, Thickening, Hair Gro... & All Natural Hair care products & Natural Hair Growth \& Beard Care Products & Protect Your Hair and Beard With All Natural Oils \\
				\midrule
				Royal 310DX Thermal Print Electronic Cash Register & Affordable Reliable Cash Management from Royal & Royal Cash Registers - Retail \& Event Supplies & Secure your cash with Royal Cash Registers \\
				\midrule
				Blue Copper 5 Anti-Aging Body Lift, Pregnancy Stretch Marks Prevention and Removal Cream 5 Oz & Blue Copper 5 Anti-Aging Products & Discover Osmotics Best Selling Products & Say Goodbye to Stretch Marks \\
				\midrule
				Cosy House Collection Twin Size Bed Sheets - Cream Bedding Set - Deep Pocket - Extra Soft Luxury... & Soft \& Hypoallergenic Twin Sheets & Luxury Twin Sheets - These Will Change Your Life. & Soft Luxury Sheets - These Will Change Your Life. \\
				\midrule
				Carson Dellosa | Valentine's Day Heart Stickers | 1-inch x 1-inch, 216ct & Share the Love with this Classroom Decor & Valentine's Day Celebrations & Show your Valentine some love this Valentine's Day \\
				\midrule
				Canon GI-20 PGBK Ink Bottle, Compatible to PIXMA G6020 and G5020 MegaTank Printers & Print more for less with SuperTank printers. Canon & All-in-one solution for professional grade prints. & All-in-one solution for professional grade prints. \\
				\midrule
				LABILUS iPhone Xs MAX case, (Rugged Armor Series) TPU Soft Stripe Designed Protective Cover Case... & 360\textdegree{} Protection Heavy Duty for iPhone Xs Max & Rugged Protective Case for iPhone Xs MAX & Rugged Armor Protective Case for iPhone Xs Max \\
				\midrule
				Biscotti Cookie Gift Basket, Gourmet Gift Basket, Delicious Biscotti Artfully Decorated 18 Count... & Valentines Gifts & Gourmet Chocolate Gift Baskets & Gourmet Holiday Gift Baskets \\
				\midrule
				Le Angelique Tapered Curling Iron Wand with Glove And 2 Clips - 3/4 to 1 Inch (18-25mm) Conical ... & Tapered Curling Wands with Glove and 2 Clips & Le Angelique Tapered Curling Iron Wand & Le Angelique Tapered Curling Iron Wand \\
				\midrule
				JUNK Brands London Fog-BBL London Fog Big Bang Lite Headband & Headbands for Every Adventure & Headbands for Every Adventure & BBL Headbands for Adventure \\
				\midrule
				Jump\&Go Portable Car Battery Jump Starter set -16,000mAh, 600A Peak, Mini Automotive Power Boost... & Portable Jump and go Jumpstarter & Jump and Go Portable Car Jump Starter & Jump and Go Portable Jump Starters \\
				\midrule
				decanit Silver Metal Thin Edge 5x7 Picture Frames, Silver Thin Profile Photo Frames 5 by 7 Inch,... & Life-Style-Living & Metal Thin Edge Picture Frames & Thin Edge Picture Frames \\
				\midrule
				CARSYS Coating Thickness Gauge DPM-816 Extended Range Precision Probe Fe/NFe Paint Meter for Car... & Coating Thickness Gauges & Coating Thickness Gauge - Range Precision & Coating Thickness Gauge \\
				\midrule
				Rich \& Creamy Buttermilk Syrup Original Flavor by Uncle Bob's Butter Country 16 fl oz/1 Pack & Fresh and Premium Buttermilk Syrup & Rich and Creamy Buttermilk Syrup & Rich and Creamy Buttermilk Syrup - Taste Great \\
				\midrule
				Sesame Street Ernie Face Tee Funny Humor Pun Adult Mens Graphic T-Shirt Apparel (Small), Orange & Sesame Street Tees for Adults & Sesame Street Men's Shirts & Sesame Street Men's Shirts \\
				\midrule
				Agvee Unbreakable End Tip [3 Pack 6ft] 4A Heavy Duty USB Phone Charger Cable, Durable Charging for iPhone 11 Pro Max X XS XR, i-Phone 10x 10xs ...... & AGVEE Fast Lightning Charging Cable for iPhone & AGVEE Fast iPhone 11 X 10s 10s XR Cable & AGVEE Heavy Duty iPhone 11 Xs XS XR Cable \\
				\bottomrule
			\end{tabular}
		\end{scriptsize}
	\end{center}
	\vskip -0.1in
	\caption{Some samples of model generated headlines from subsets rated 3, 2 and 1. The frequency of headlines is not indicative of true distribution of headline quality.}
	\label{table:sample_headlines}
\end{table*}

We also conducted large scale crowd-sourced evaluation studies of the headlines with over 150,000 judgments. All headlines are shuffled and each headline is rated by 3 random and double-blind crowd-sourced auditors. The quality is judged on a 3-point scale of [1. Incorrect or Irrelevant, 2. Correct, 3. Correct and Attractive] and we use the mode of the 3 judgments.

In this double-blind audit, the auditors were not aware of the source of the headlines and we were not aware of the identity or demographics of any auditor. More details about the workforce may be found in the platform documentation \citep{ground_truth_2021}. In order to determine the compensation for the crowd-sourced workers, we used the guideline provided by the crowd-sourcing platform to ``choose a price consistent with the approximate time it takes to complete a task" (Visible in the Console while creating the \citet{labeling_2021} job). We thus first conducted an internal audit by volunteers across our organization to determine the time required to complete the task (average 21.59s) and then used the remuneration recommended for the corresponding time range (\$0.12 for 20s - 22s).

Table \ref{table:quality_scores} summarizes the quality audits. The SC-biLSTM model performed worse compared to human-submitted headlines. The proposed SC-MLM model achieves the highest average rating and the most number of perfectly rated headlines. Using just a single product does produce correct headlines with 8\% faster inference latency but fails to produce attractive headlines due to lack of input from multiple products.

We also conducted  Grammar specific audits (N $\approx$ 10000) in which the grammar of the headlines is judged independently. 98.13\% of SC-MLM and  98.12\% of MLM generated headlines were judged to have correct grammar against 93.14\% of human submitted headlines.

Table \ref{table:sample_headlines} shows a sample of headlines for campaigns in the blind test-set. Excessive keyword stuffing in source product titles does hamper headline quality at times and post-filtering using beam search score helps to filter them out. We do observe cases where both the models generate the same headline. This is an artifact of the fact that both the models share the first 15 epochs. The SC-MLM model generates more descriptive headlines and both models are able to abstract the product qualities.

\section{Conclusion}
Ad headline generation is a difficult problem owing to the varying nature of retail product attributes. A lot of historical methods focus on template based creation of ad headlines that are not very expressive.

We demonstrated a new NLG based method to generate headlines for multiple products. Our method achieves highest score in overlap metrics, quality audits and grammar audits compared to the baselines and human-submitted headlines.

Masked Language Models were relatively unexplored for ad headline generation and we were able to demonstrate their utility. We further extended the performance of the model by using Reinforcement Learning. The method only changes the training procedure without impacting inference latency. Thus, our work contributes to both SOTA and practical business applications.

The approach can also be used for any other NLG task.



\newpage

\bibliography{anthology,custom}

\begin{thebibliography}{25}
\expandafter\ifx\csname natexlab\endcsname\relax\def\natexlab#1{#1}\fi

\bibitem[{Bahdanau et~al.(2017)Bahdanau, Brakel, Xu, Goyal, Lowe, Pineau,
  Courville, and Bengio}]{bahdanau_actor-critic_2017}
Dzmitry Bahdanau, Philemon Brakel, Kelvin Xu, Anirudh Goyal, Ryan Lowe, Joelle
  Pineau, Aaron Courville, and Yoshua Bengio. 2017.
\newblock \href {http://arxiv.org/abs/1607.07086} {An {Actor}-{Critic}
  {Algorithm} for {Sequence} {Prediction}}.
\newblock \emph{arXiv:1607.07086 [cs]}.
\newblock ArXiv: 1607.07086.

\bibitem[{Devlin et~al.(2019)Devlin, Chang, Lee, and
  Toutanova}]{devlin_bert_2019}
Jacob Devlin, Ming-Wei Chang, Kenton Lee, and Kristina Toutanova. 2019.
\newblock \href {http://arxiv.org/abs/1810.04805} {{BERT}: {Pre}-training of
  {Deep} {Bidirectional} {Transformers} for {Language} {Understanding}}.
\newblock \emph{arXiv:1810.04805 [cs]}.
\newblock ArXiv: 1810.04805.

\bibitem[{Dong et~al.(2019)Dong, Yang, Wang, Wei, Liu, Wang, Gao, Zhou, and
  Hon}]{dong_unified_2019}
Li~Dong, Nan Yang, Wenhui Wang, Furu Wei, Xiaodong Liu, Yu~Wang, Jianfeng Gao,
  Ming Zhou, and Hsiao-Wuen Hon. 2019.
\newblock \href {http://arxiv.org/abs/1905.03197} {Unified {Language} {Model}
  {Pre}-training for {Natural} {Language} {Understanding} and {Generation}}.
\newblock \emph{arXiv:1905.03197 [cs]}.
\newblock ArXiv: 1905.03197.

\bibitem[{Ground Truth(2021)}]{ground_truth_2021}
\newblock Using {MTurk} with {Ground} {Truth}. 2021.
\newblock [link].

\bibitem[{Hughes et~al.(2019)Hughes, Chang, and Zhang}]{hughes_generating_2019}
J.~Weston Hughes, Keng-hao Chang, and Ruofei Zhang. 2019.
\newblock \href {https://doi.org/10.1145/3292500.3330754} {Generating {Better}
  {Search} {Engine} {Text} {Advertisements} with {Deep} {Reinforcement}
  {Learning}}.
\newblock In \emph{Proceedings of the 25th {ACM} {SIGKDD} {International}
  {Conference} on {Knowledge} {Discovery} \& {Data} {Mining}}, {KDD} '19, pages
  2269--2277, Anchorage, AK, USA. Association for Computing Machinery.

\bibitem[{Joulin et~al.(2016)Joulin, Grave, Bojanowski, and
  Mikolov}]{joulin_bag_2016}
Armand Joulin, Edouard Grave, Piotr Bojanowski, and Tomas Mikolov. 2016.
\newblock Bag of tricks for efficient text classification.
\newblock \emph{arXiv preprint arXiv:1607.01759}.

\bibitem[{Kingma and Ba(2014)}]{adam}
Diederik Kingma and Jimmy Ba. 2014.
\newblock Adam: A method for stochastic optimization.
\newblock \emph{International Conference on Learning Representations}.

\bibitem[{Labeling(2021)}]{labeling_2021}
Ground~Truth Labeling. 2021.
\newblock Create a labeling job.

\bibitem[{Lewis et~al.(2019)Lewis, Liu, Goyal, Ghazvininejad, Mohamed, Levy,
  Stoyanov, and Zettlemoyer}]{lewis_bart_2019}
Mike Lewis, Yinhan Liu, Naman Goyal, Marjan Ghazvininejad, Abdelrahman Mohamed,
  Omer Levy, Ves Stoyanov, and Luke Zettlemoyer. 2019.
\newblock \href {http://arxiv.org/abs/1910.13461} {{BART}: {Denoising}
  {Sequence}-to-{Sequence} {Pre}-training for {Natural} {Language}
  {Generation}, {Translation}, and {Comprehension}}.
\newblock \emph{arXiv:1910.13461 [cs, stat]}.
\newblock ArXiv: 1910.13461.

\bibitem[{Lin(2004)}]{lin_rouge_2004}
Chin-Yew Lin. 2004.
\newblock \href {https://www.aclweb.org/anthology/W04-1013} {{ROUGE}: {A}
  {Package} for {Automatic} {Evaluation} of {Summaries}}.
\newblock In \emph{Text {Summarization} {Branches} {Out}}, pages 74--81,
  Barcelona, Spain. Association for Computational Linguistics.

\bibitem[{Papineni et~al.(2002)Papineni, Roukos, Ward, and
  Zhu}]{papineni_bleu_2002}
Kishore Papineni, Salim Roukos, Todd Ward, and Wei-Jing Zhu. 2002.
\newblock \href {https://doi.org/10.3115/1073083.1073135} {Bleu: a method for
  automatic evaluation of machine translation}.
\newblock In \emph{Proceedings of the 40th {Annual} {Meeting} of the
  {Association} for {Computational} {Linguistics}}, pages 311--318,
  Philadelphia, Pennsylvania, USA. Association for Computational Linguistics.

\bibitem[{policies()}]{amazon_advertising_policies_sponsored_link}
\newblock Sponsored {Advertising} policies.
\newblock [link].

\bibitem[{Radford et~al.(2019)Radford, Wu, Child, Luan, Amodei, and
  Sutskever}]{radford_language_2019}
Alec Radford, Jeffrey Wu, Rewon Child, David Luan, Dario Amodei, and Ilya
  Sutskever. 2019.
\newblock Language models are unsupervised multitask learners.
\newblock \emph{OpenAI Blog}, 1(8):9.

\bibitem[{Ramachandran et~al.(2016)Ramachandran, Liu, and
  Le}]{ramachandran_unsupervised_2016}
Prajit Ramachandran, Peter~J. Liu, and Quoc~V. Le. 2016.
\newblock Unsupervised pretraining for sequence to sequence learning.
\newblock \emph{arXiv preprint arXiv:1611.02683}.

\bibitem[{Ranzato et~al.(2016)Ranzato, Chopra, Auli, and
  Zaremba}]{ranzato_sequence_2016}
Marc'Aurelio Ranzato, Sumit Chopra, Michael Auli, and Wojciech Zaremba. 2016.
\newblock \href {http://arxiv.org/abs/1511.06732} {Sequence {Level} {Training}
  with {Recurrent} {Neural} {Networks}}.
\newblock \emph{arXiv:1511.06732 [cs]}.
\newblock ArXiv: 1511.06732.

\bibitem[{Rennie et~al.(2017)Rennie, Marcheret, Mroueh, Ross, and
  Goel}]{rennie_self-critical_2017}
Steven~J. Rennie, Etienne Marcheret, Youssef Mroueh, Jarret Ross, and Vaibhava
  Goel. 2017.
\newblock \href {http://arxiv.org/abs/1612.00563} {Self-critical {Sequence}
  {Training} for {Image} {Captioning}}.
\newblock \emph{arXiv:1612.00563 [cs]}.
\newblock ArXiv: 1612.00563.

\bibitem[{Rothe et~al.(2019)Rothe, Narayan, and
  Severyn}]{rothe_leveraging_2019}
Sascha Rothe, Shashi Narayan, and Aliaksei Severyn. 2019.
\newblock \href {http://arxiv.org/abs/1907.12461} {Leveraging {Pre}-trained
  {Checkpoints} for {Sequence} {Generation} {Tasks}}.
\newblock \emph{arXiv:1907.12461 [cs]}.
\newblock ArXiv: 1907.12461.

\bibitem[{See et~al.(2017)See, Liu, and Manning}]{see_get_2017}
Abigail See, Peter~J. Liu, and Christopher~D. Manning. 2017.
\newblock \href {https://doi.org/10.18653/v1/P17-1099} {Get {To} {The} {Point}:
  {Summarization} with {Pointer}-{Generator} {Networks}}.
\newblock In \emph{Proceedings of the 55th {Annual} {Meeting} of the
  {Association} for {Computational} {Linguistics} ({Volume} 1: {Long}
  {Papers})}, pages 1073--1083, Vancouver, Canada. Association for
  Computational Linguistics.

\bibitem[{Sharma et~al.(2017)Sharma, Asri, Schulz, and
  Zumer}]{sharma_relevance_2017}
Shikhar Sharma, Layla~El Asri, Hannes Schulz, and Jeremie Zumer. 2017.
\newblock \href {http://arxiv.org/abs/1706.09799} {Relevance of {Unsupervised}
  {Metrics} in {Task}-{Oriented} {Dialogue} for {Evaluating} {Natural}
  {Language} {Generation}}.
\newblock \emph{arXiv:1706.09799 [cs]}.
\newblock ArXiv: 1706.09799.

\bibitem[{Vaswani et~al.(2017)Vaswani, Shazeer, Parmar, Uszkoreit, Jones,
  Gomez, Kaiser, and Polosukhin}]{vaswani_attention_2017}
Ashish Vaswani, Noam Shazeer, Niki Parmar, Jakob Uszkoreit, Llion Jones,
  Aidan~N Gomez, Łukasz Kaiser, and Illia Polosukhin. 2017.
\newblock \href
  {http://papers.nips.cc/paper/7181-attention-is-all-you-need.pdf} {Attention
  is {All} you {Need}}.
\newblock In I.~Guyon, U.~V. Luxburg, S.~Bengio, H.~Wallach, R.~Fergus,
  S.~Vishwanathan, and R.~Garnett, editors, \emph{Advances in {Neural}
  {Information} {Processing} {Systems} 30}, pages 5998--6008. Curran
  Associates, Inc.

\bibitem[{Williams(1992)}]{williams_simple_1992}
Ronald~J. Williams. 1992.
\newblock \href {https://doi.org/10.1007/BF00992696} {Simple statistical
  gradient-following algorithms for connectionist reinforcement learning}.
\newblock \emph{Machine Learning}, 8(3):229--256.

\bibitem[{Wolf et~al.(2020)Wolf, Debut, Sanh, Chaumond, Delangue, Moi, Cistac,
  Rault, Louf, Funtowicz, and Brew}]{wolf_huggingfaces_2020}
Thomas Wolf, Lysandre Debut, Victor Sanh, Julien Chaumond, Clement Delangue,
  Anthony Moi, Pierric Cistac, Tim Rault, Rémi Louf, Morgan Funtowicz, and
  Jamie Brew. 2020.
\newblock \href {http://arxiv.org/abs/1910.03771} {{HuggingFace}'s
  {Transformers}: {State}-of-the-art {Natural} {Language} {Processing}}.
\newblock \emph{arXiv:1910.03771 [cs]}.
\newblock ArXiv: 1910.03771.

\bibitem[{Wu et~al.(2016)Wu, Schuster, Chen, Le, Norouzi, Macherey, Krikun,
  Cao, Gao, Macherey, Klingner, Shah, Johnson, Liu, Kaiser, Gouws, Kato, Kudo,
  Kazawa, Stevens, Kurian, Patil, Wang, Young, Smith, Riesa, Rudnick, Vinyals,
  Corrado, Hughes, and Dean}]{wu_googles_2016}
Yonghui Wu, Mike Schuster, Zhifeng Chen, Quoc~V. Le, Mohammad Norouzi, Wolfgang
  Macherey, Maxim Krikun, Yuan Cao, Qin Gao, Klaus Macherey, Jeff Klingner,
  Apurva Shah, Melvin Johnson, Xiaobing Liu, Łukasz Kaiser, Stephan Gouws,
  Yoshikiyo Kato, Taku Kudo, Hideto Kazawa, Keith Stevens, George Kurian,
  Nishant Patil, Wei Wang, Cliff Young, Jason Smith, Jason Riesa, Alex Rudnick,
  Oriol Vinyals, Greg Corrado, Macduff Hughes, and Jeffrey Dean. 2016.
\newblock \href {http://arxiv.org/abs/1609.08144} {Google's {Neural} {Machine}
  {Translation} {System}: {Bridging} the {Gap} between {Human} and {Machine}
  {Translation}}.
\newblock \emph{arXiv:1609.08144 [cs]}.
\newblock ArXiv: 1609.08144.

\bibitem[{Xu et~al.(2019)Xu, Wu, Madotto, and Fung}]{xu_clickbait_2019}
Peng Xu, Chien-Sheng Wu, Andrea Madotto, and Pascale Fung. 2019.
\newblock \href {https://doi.org/10.18653/v1/D19-1303} {Clickbait?
  {Sensational} {Headline} {Generation} with {Auto}-tuned {Reinforcement}
  {Learning}}.
\newblock In \emph{Proceedings of the 2019 {Conference} on {Empirical}
  {Methods} in {Natural} {Language} {Processing} and the 9th {International}
  {Joint} {Conference} on {Natural} {Language} {Processing}
  ({EMNLP}-{IJCNLP})}, pages 3065--3075, Hong Kong, China. Association for
  Computational Linguistics.

\bibitem[{Zhu et~al.(2020)Zhu, Xia, Wu, He, Qin, Zhou, Li, and
  Liu}]{zhu_incorporating_2020}
Jinhua Zhu, Yingce Xia, Lijun Wu, Di~He, Tao Qin, Wengang Zhou, Houqiang Li,
  and Tie-Yan Liu. 2020.
\newblock \href {http://arxiv.org/abs/2002.06823} {Incorporating {BERT} into
  {Neural} {Machine} {Translation}}.
\newblock \emph{arXiv:2002.06823 [cs]}.
\newblock ArXiv: 2002.06823.

\end{thebibliography}
\bibliographystyle{acl_natbib}

\appendix

%

\end{document}